\documentclass{sig-alternate-2013}

\usepackage{quoting}
\quotingsetup{vskip=3pt,leftmargin=0pt,rightmargin=0pt,indentfirst=false}
\usepackage{xspace}
\usepackage{times}
\usepackage{graphicx}
\usepackage{latexsym}
\usepackage{hyperref}
\usepackage{url}
\usepackage[normalem]{ulem}

\newcommand{\eg}{e.g.\xspace}

\newcommand{\bleu}{\textsc{BLEU}\xspace}
\newcommand{\rouge}{\textsc{ROUGE}\xspace}

\permission{\copyright 2017 International World Wide Web Conference Committee \\ (IW3C2), published under Creative Commons CC BY 4.0 License.}
\conferenceinfo{WWW'17 Companion,}{April 3--7, 2017, Perth, Australia.}
\copyrightetc{ACM \the\acmcopyr}
\crdata{978-1-4503-4914-7/17/04. \\
http://dx.doi.org/10.1145/3041021.3054264 \\
\includegraphics{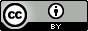}
}

\begin{document}

\title{Post-edit Analysis of Collective Biography Generation}

\numberofauthors{1} 
\author{
\alignauthor
Bo Han~~~Will Radford~~~Ana{\" i}s Cadilhac~~~Art Harol~~~Andrew Chisholm~~~Ben Hachey\\
    \affaddr{Hugo.ai}\\
    \affaddr{58-62 Kippax Street}\\
    \affaddr{Surry Hills, Australia}\\
    \email{\{bhan,wradford,acadilhac,aharol,achisholm,bhachey\}@hugo.ai}
}

\additionalauthors{Additional authors: John Smith (The Th{\o}rv{\"a}ld Group,
email: {\texttt{jsmith@affiliation.org}}) and Julius P.~Kumquat
(The Kumquat Consortium, email: {\texttt{jpkumquat@consortium.net}}).}
\date{30 July 1999}

\maketitle
\begin{abstract}
Text generation is increasingly common but often requires manual post-editing where high precision is critical to end users.
However, manual editing is expensive so we want to ensure this effort is focused on high-value tasks.
And we want to maintain stylistic consistency, a particular challenge in crowd settings.
We present a case study, analysing human post-editing in the context of a template-based biography generation system.
An edit flow visualisation combined with manual characterisation of edits helps identify and prioritise work for improving end-to-end efficiency and accuracy.
\end{abstract}


\section{Introduction}

Natural language generation applications often use human post-editing to ensure quality of final output.
Understanding the kind and amount of manual effort is critical to prioritising how generation can be improved and what editorial guidelines should be implemented.
Recent work also suggests that some post-editing can be automated, with substantial improvements in a machine translation shared task (+5.5 \bleu score) \cite{wmt16ape}.

Post-edit analysis also has a long history in the evaluation of text generation tasks, in particular machine translation \cite{Aziz_etal:2012,scarton:2015:SRW}.
The closest work to ours \cite{sripada-ijcai05-evaluating} is a post-edit evaluation of 2,728 pairs of system-edited texts from a system for generating weather forecasts, reporting a range of edits including individual style preferences as well as corrections.
We build on this work to perform a post-edit analysis complementary to existing evaluation metrics like \rouge and \bleu.
We introduce a visualisation that helps identify and prioritise improvements, and present a case study using data from a commercial biography generation system, reporting a detailed analysis of edit flow and characteristic edit actions.

\newpage
\section{System Overview}
\label{sec:system}

Our case study evaluates an collective biography generation system that is part of a larger commercial tool for finding and summarising information about a person on the web.
Input comprises facts and events derived from social and news sources (\eg, name, affiliations, education, skills, investments).
Facts and events are obtained from a search tool, which guides internal crowd users through the search process to select a range of relevant and diverse sources.
The core generation capability uses 26 templates to produce an average of 2.8 suggested sentences per biography, \eg:
\begin{quoting}
    \small
    \emph{\textbf{Suggested:} Philip has been the Co-Founder at High Fidelity, Inc.\ since January 2013 and an Investor at Milk, Sunglass.io, Akili Interactive, Crowdfunder and more, and specializes in virtual worlds, start-ups and software development. He received a BS degree in Physics from The University of California, San Diego in 1992. Philip frequently tweets the hashtags \#vr, \#highfidelity and \#avatars. Philip recently interacted with @id\_aa\_carmack and @inc on Twitter. Philip's favorite movie is Meditate and Destroy. Philip is a member of Facebook groups ``Virtual Blogging'', ``San Francisco Bay Area Free Yoga and Meditation Events'' and ``BVHS Class of 1986''.}
\end{quoting}
Internal crowd users select, edit and add suggested sentences, \eg:
\begin{quoting}
    \small
    \emph{\textbf{Selected:} Philip has been the Co-Founder at High Fidelity, Inc.\ since January 2013 and an Investor at Milk, Sunglass.io, Akili Interactive, Crowdfunder and more, and specializes in virtual worlds, start-ups and software development. \textbf{Edited:} He `received'$\mapsto$`holds' a BS degree in Physics from The University of California, San Diego in 1992. \textbf{Selected:} Philip frequently tweets the hashtags \#vr, \#highfidelity and \#avatars. \textbf{Not Selected:} \sout{Philip recently interacted with @id\_aa\_carmack and @inc on Twitter.} \textbf{Not Selected:} \sout{Philip's favorite movie is Meditate and Destroy. Philip is a member of Facebook groups ``Virtual Blogging'', ``San Francisco Bay Area Free Yoga and Meditation Events'' and ``BVHS Class of 1986''.} \textbf{New:} He is married to Yvette Forte Rosedale. \textbf{New:} In 2007, Philip was listed in Time Magazine's 100 Most Influential People in The World.}
\end{quoting}
The result is a concise and informative text for end users, \eg:
\begin{quoting}
    \small
    \emph{\textbf{Final:} Philip has been the Co-Founder at High Fidelity, Inc.\ since January 2013 and an Investor at Milk, Sunglass.io, Akili Interactive, Crowdfunder and more, and specializes in virtual worlds, start-ups and software development. He received a BS degree in Physics from The University of California, San Diego in 1992.  Philip frequently tweets the hashtags \#vr, \#highfidelity and \#avatars. He is married to Yvette Forte Rosedale. In 2007, Philip was listed in Time Magazine's 100 Most Influential People in The World.}
\end{quoting}

The goal of the analysis here is twofold: (1) identify areas for improving system coverage and precision to focus human effort on high-value cognitively demanding tasks and improve end-to-end efficiency; (2) identify strategies for improving consistency and quality of biographies produced from noisy web data.
The analysis uses 10,320 final biographies for which we also have the original suggested sentences from the generation system.
This allows a unique and detailed characterisation of collective biography generation, tracking the editorial process from automatically generated sentences to final sentences after human post-editing.


\newpage
\section{Visualising Edit Flow}
\label{sec:setting}

\textbf{Sentence alignment}
We first align suggested with final sentences 
using a minimum Levenshtein ratio of 0.8, defined as $(l_{(a+b)}-d(a,b))/l_{(a+b)}$ where $l_{(a+b)}$ is the combined length of the input strings and $d(a,b)$ is the Levenshtein edit distance between the input strings.
Suggested biographies have a total 28,842 sentences and 525,555 tokens, while final biographies have a total of 33,238 sentences and 580,521 tokens.
So, final biographies tend to have more sentences (3.2 versus 2.8), but each sentence is shorter on average (17.5 versus 18.2).

\begin{figure}
  \centering
  \includegraphics[width=\columnwidth]{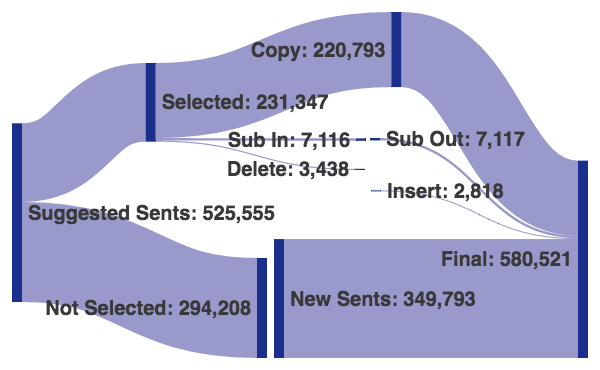}
  \caption{Flow of tokens through the post-editing process.}
  \label{sankey}
\end{figure}

\textbf{Edit flow}
Figure \ref{sankey} visualises the flow of tokens from suggestions through to the final biography using a Sankey diagram.
The total number of tokens in suggested sentences breaks down into two categories: 231,347 (44\%) are selected by users and 220,793 (56\%) are not selected.
Looking at tokens in final sentences: 349,793 (60\%) are new sentences written by human editors.
This suggests that work on improving end-to-end efficiency should prioritise extending fact collection and generation to provide more useful suggestions (more detail in Section \ref{sec:characterising_edits}).
Trimming unused suggestions is less valuable as ignoring these is relatively easy.

\textbf{Token alignment}
We refine the edit flow analysis by identifying phrase substitutions within the selected sentences aligned in the first step. 
We leverage the standard diff library in Python to collect token-level edit operations.
This provides more detailed phrase-level substitution, deletion and insertion statistics corresponding to the middle of the edit flow diagram.
Interestingly, we find that the vast majority (95\%) of tokens in aligned sentences are copied directly from selected to final sentences. 
Phrase edits collectively represent only 5\% of tokens in selected sentences, suggesting that generation obtains high accuracy where facts and templates are available.


\section{Characterising Edits}
\label{sec:characterising_edits}

\textbf{Selected}
From token alignments in selected sentences, we observe a total of 1,833 substitution, 621 deletion and 468 insertion phrase pairs and 4,568, 1,343 and 1,284 instances respectively. 
Overall, selection edits tend to result in shorter sentences with an average of 1.8 tokens before editing compared to 1.6 post-editing.
Table \ref{tab:frequent_edits} lists some of the most common edit actions.
We observe that the top edits are generally grammatical (\eg, `the Engineer'$\mapsto$`an Engineer') or stylistic (\eg, `received a BFA'$\mapsto$`holds a BFA').
Other edits address capitalisation consistency or errors from collected facts (\eg, `The University'$\mapsto$`the University', `microsoft office'$\mapsto$`Microsoft Office').

\begin{table}
    \centering
    \begin{tabular}{rr@{$\mapsto$}lrr}
        \hline
        Rank & \multicolumn{2}{c}{Edit} & \multicolumn{2}{c}{\# Instances} \\
        \hline\hline
        1 & \texttt{`of'} & \texttt{`at'} & 320 & (4.4\%) \\
        2 & \texttt{`'} & \texttt{`the'} & 272 & (3.8\%) \\
        3 & \texttt{`'} & \texttt{`also'} & 233 & (3.2\%) \\
        4 & \texttt{`a'} & \texttt{`the'} & 166 & (2.3\%) \\
        5 & \texttt{`with'} & \texttt{`in'} & 130 & (1.8\%) \\
        19 & \texttt{`The'} & \texttt{`the'} & 39 & (0.5\%) \\
        20 & \texttt{`has been'} & \texttt{`is'} & 34 & (0.5\%) \\
        27 & \texttt{`has'} & \texttt{`have'} & 25 & (0.3\%) \\
        30 & \texttt{`the'} & \texttt{`an'} & 23 & (0.3\%) \\
        38 & \texttt{`received'} & \texttt{`holds'} & 20 & (0.3\%) \\
        \hline
    \end{tabular}
    \caption{Example edit actions.}
    \label{tab:frequent_edits}
\end{table}


\textbf{Not Selected}
Ignoring the sentence alignment threshold, we find that 67\% of suggested sentences that were not selected have some overlap with sentences in the final biography.
The remaining 33\% were not directly useful for their biographies.
By inspection of a random 100 of these, we observe that most talk about hobbies (\eg, interests, travel, books), social media interactions, or a monolingual person's sole language.


\textbf{New Sentences}
Looking from the other side, we find that 53\% of new sentences have some overlap with suggestions.
These comprise sentences where editors deleted/introduced many tokens, or split/merged suggested sentences.
The remaining 47\% of new sentences are due primarily to coverage issues in search or fact collection feeding into generation.

%

\section{Discussion}

We introduced an edit flow visualisation for analysis of human post-editing of automatically generated text.
This helps to identify and prioritise work for improving end-to-end efficiency.
It is encouraging that most new sentences include suggested content, but these sentences still represent significant human cost.
This suggests that end-to-end efficiency can be improved most by expanding generation templates to provide editors more style and content options.
The next priority is then to automate post-edits where possible, learnt from data here or external resources.
Another issue discovered during analysis is the lack of clear editorial policy, with users sometimes making conflicting edits.
Editorial style policy is challenging to implement with crowd-sourced editing, but we expect that gradual introduction of high-precision rules should actively guide editors toward a more consistent style.
In current work, we are implementing the changes suggested above, developing a virtuous circle to improve both efficiency and quality over time.

\bibliographystyle{abbrv}
\bibliography{www17_summ}

\end{document}